\pdfoutput=1

\documentclass[11pt]{article}

\usepackage[final]{acl}
\usepackage{amsmath}
\usepackage{tcolorbox}

\usepackage{times}
\usepackage{latexsym}
\usepackage{enumitem}

\usepackage[T1]{fontenc}

\usepackage[utf8]{inputenc}

\usepackage{microtype}

\usepackage{inconsolata}

\usepackage{graphicx}

\usepackage{subcaption}

\title{RALS: Resources and Baselines for \\Romanian Automatic Lexical Simplification}

 \author{Fabian Anghel \and Petru Theodor Cristea \and Claudiu Creangă \and Sergiu Nisioi\thanks{Corresponding author.} \\
         Human Language Technologies Research Center \\
         Faculty of Mathematics and Computer Science \\ 
         University of Bucharest \\ 
         \texttt{sergiu.nisioi@unibuc.ro}
         }

\begin{document}
\maketitle
\begin{abstract}
We introduce the first dataset that jointly covers both lexical complexity prediction (LCP) annotations and lexical simplification (LS) for Romanian, along with a comparison of lexical simplification approaches. We propose a methodology for ordering simplification suggestions using a pairwise ranking approximation method, arranging candidates from simple to complex based on a separate set of human judgments. In addition, we provide human lexical complexity annotations for 3,921 word samples in context. Finally, we explore several novel pipelines for complexity prediction and simplification and present the first text simplification system for Romanian.\footnote{
\url{https://github.com/senisioi/RALS}
}
\end{abstract}

\section{Introduction}
Text simplification is the process of transforming texts into variants that are simpler to understand by larger audiences or easier to process by existing NLP systems. Such initiatives promote literacy, facilitate effective communication, and enable equal access to information for individuals with diverse reading abilities or special needs \cite{zilio-etal-2020-lexical,vstajner2021automatic,gooding2022ethical}. These outcomes have broad-reaching advantages in sectors such as education, healthcare, legal documentation, government communication, online content, and beyond, ultimately enhancing social inclusivity and empowerment. 

Unlike sentence simplification, which is typically modeled as a monolingual machine translation task \cite{specia2010translating,nisioi-etal-2017-exploring,dou2023automatic}, lexical simplification (LS) is specifically targeting particular lexical items to better control text generation and evaluation \cite{devlin1998use,carroll1998practical,de2010text,glavas-stajner-2015-simplifying,lee-yeung-2018-personalizing,sheang-etal-2022-controllable,gooding-tragut-2022-one}.
In this way, the output can be guided towards specific target groups such as children or readers with different degrees of literacy.

End-to-end lexical simplification is typically divided into two equally challenging subtasks: \textbf{a) lexical complexity prediction (LCP)}, which assigns complexity scores to words \cite{yimam-etal-2018-report} and \textbf{b) lexical simplification}, which suggests simpler replacement words guided by LCP scores through candidate retrieval and re-ranking \cite{north2023lexical}.

The series of workshops on lexical complexity prediction and lexical simplification \cite{specia2012semeval,paetzold-specia-2016-semeval,yimam-etal-2018-report,shardlow2021semeval,saggion2023findings,shardlow2024bea,tsar-2024-text} along with their shared tasks have nurtured a growing international interest in multilingual simplification. Several resources have been independently developed and used in these tasks that cover well-resourced languages such as Spanish \cite{ferres-saggion-2022-alexsis,alarcon2023easier,stajner-etal-2023-less}, German \cite{ebling2022automatic}, Dutch \cite{hobo-etal-2023-geen}, Portuguese \cite{hartmann2020adaptaccao,north2022alexsis}, Japanese \cite{kajiwara2015evaluation,kodaira-etal-2016-controlled,ide2023}, Chinese \cite{qiang2021chinese}, and French \cite{billami2018resyf,pintard-francois-2020-combining}.
Despite its potential broad impact, this task has received relatively little attention for medium and low-resource languages, where Italian, Catalan, Sinhala, or Filipino \cite{shardlow-etal-2024-extensible}, have remained relatively underexplored, and Eastern Romance languages such as Romanian are completely absent from landscape text simplification research.
This scarcity of resources is not unique to Romanian \cite{codruț2024rodia}; many other languages worldwide face similar challenges due to their limited visibility on the global linguistic stage.

In this paper, we address the twin challenges of creating annotated datasets for lexical complexity prediction and lexical simplification in Romanian, a language currently lacking resources in this domain. Furthermore, we close this resource gap by constructing several end-to-end lexical simplification models specifically adapted to the particularities of Romanian.

\section{Challenges and Related Work}

Given the extremely low-resource setting, training end-to-end systems for Romanian comes with several challenges that cover both data construction and training techniques. To have comparable results across languages, LCP data should ideally be annotated with similar guidelines, for similar target groups, and comparable text genres across languages. One such attempt is the MultiLS dataset \cite{shardlow2024readi} developed for the 2024 Shared Task \cite{shardlow2024bea} which covers ten languages with similar methodological annotations, even though genres and annotator target groups differ across languages. Our data construction follows a multi-step approach involving human translation, machine translation, and multiple types of manual annotation, including self-assessment for lexical complexity prediction (LCP) and pairwise annotation for ranking simplification candidates.



The 2024 Shared Task \cite{shardlow2024bea} showed that training systems from scratch on limited or synthetically generated data \cite{sastre-etal-2024-retuyt} yields poorer results (Pearson $r \approx 0.4$) than zero-shot prompting with GPT-4 ($r \approx 0.6$) \cite{enomoto-etal-2024-tmu}. However, the better-performing approach should not be considered a \textit{silver bullet}, as prompting proprietary LLM systems to provide complexity score assessments carries practical risks, including privacy leaks, hallucinations, and uncontrolled output variability \cite{yao2024survey,AllenZhu-icml2024-tutorial}.

An alternative approach that does not use LLMs was tested by \citet{cristeaNisioi24}, who built cross-lingual LCP predictors using machine translations (MT) into English and back-translations; however, their experiments produced weak results ($r \approx 0.3$). MT has significant pitfalls: words that are easy in one language might not be easy in another, translations are rarely done word-by-word, and \textit{translationese} is a language variety and lect with its own particularities \cite{blum1978universals,rabinovich2016similarities}. In our work, we incorporate both carefully curated translations and words sampled from original texts written in Romanian. We create a high-quality set of data. Each sentence and word is carefully chosen to have both original annotations and annotations comparable with those in English.

Regarding lexical simplification, rule-based methods consisting of different pipelines such as word-sense disambiguation, synonym reranking, and morphological operations \cite{paetzold-specia-2015-lexenstein,ferres2017adaptable} are less prevalent and are considered weaker than neural models because they depend on high-quality linguistic resources and robust pipelines.
Nevertheless, the results reported by \citet{saggion2023findings} at the TSAR Shared Task point out that some neural systems under-performed rule-based baseline.

In this work, we propose a hybrid solution that combines pre-existing synonym and morphological inflection dictionaries with a contextual embedding-based word-sense detector.

Neural network-based solutions dominate lexical simplification tasks, and we highlight two main types of systems \cite{north2024deep}: 1) masked or generative language models \cite{qiang2020lexical,qiang2021lsbert,ferres-saggion-2022-alexsis,sheang2023multilingual}, which perform word prediction and reranking, and 2) LLM-based instruction tuning \cite{baez-saggion-2023-lsllama} or prompting of closed-source systems \cite{aumiller-gertz-2022-unihd,enomoto-etal-2024-tmu}. The latter has achieved the highest performance of any LS method across languages in both the 2022 and 2024 Shared Tasks \cite{saggion2023findings,shardlow2024bea}.

Approaches based on masked language models for candidate suggestion \cite{north2024deep} are generally ineffective for low-resource languages, as they often alter sentence meaning by suggesting simplifications from semantically related categories (e.g., \textit{cat, dog, mouse; coffee, tea}). 
Furthermore, LLM prompting for candidate suggestion has difficulties in producing words in the correct inflected form and may lead to hallucinations \cite{cristeaNisioi24}. Finally, proprietary systems come with cost restrictions, lack transparency, cannot be trusted with data requiring high privacy, and there is no guarantee of result consistency.

Our work addresses several of these challenges and offers a comparative analysis of lexical simplification methods.



\begin{table*}[!htb]
\centering
\resizebox{.76\textwidth}{!}{%
\begin{tabular}{l|cccc|cccc}
 & \multicolumn{4}{c}{Sentence length} & \multicolumn{4}{c}{Complexity} \\
     & \textbf{En} & \textbf{HT} & \textbf{WT} & \textbf{RoLCP} & \textbf{En} & \textbf{HT} & \textbf{WT} & \textbf{RoLCP} \\
\hline
mean & 22.82 & 24.46 & 24.53 & 27.58 & 0.24 & 0.13 & 0.28 & 0.23 \\
std & 7.74 & 8.62 & 8.43 & 26.59 & 0.19 & 0.25 & 0.21 & 0.23 \\
min & 7 & 9 & 6 & 2 & 0.02 & 0 & 0 & 0 \\
max & 45 & 49 & 42 & 318 & 0.93 & 1 & 1 & 1 \\
no. samples & 569 & 569 & 1,765 & 1,587 & 569 & 569 & 1,765 & 1,587 \\
no. sentences & 190 & 190 & 751 & 274 & 190 & 190 & 751 & 274 \\
\end{tabular} %
}
\caption{Statistics comparing the English and Romanian Human Translation (HT) sentences (569 samples), the Romanian Word-level translation (WT) datasets (1765 samples), and the new RoLCP data (1,587 samples). In total, 3,921 annotated samples for LCP on Romanian. The annotated complexity for words occurring in original texts resemble closer the distribution of the similar word annotations in original English.
}
\label{tab:stats}
\end{table*}

\section{Data Collection}

\subsection{Lexical Complexity Dataset}
The English portion of the MultiLS dataset \cite{shardlow2024readi} contains 569 word–sentence sample pairs sourced from Wikibooks. For each sentence, three words are annotated on a scale of 1 to 5, from simple to complex, through crowd-sourcing.

We construct three Romanian subsets: 

\textbf{1. HT:} a direct counterpart, created by human translation (HT) of all 569 samples into Romanian and carefully aligning the most appropriate target word with the original English annotated equivalent; 

\textbf{2. WT:} 1,765 samples identified in the Representative Corpus of Romanian \cite{midrigan-ciochina-etal-2020-resources} using the same set of words as in the HT dataset, with the aim of testing whether sentences containing word translations (WT) offer better representativeness (see \autoref{app:wt});

\textbf{3. RoLCP:} 1,587 new samples containing words not included in HT or WT, selected based on frequency distributions, annotated in sentences drawn from diverse Romanian texts, including Wikipedia articles, popular science, literature, institutional documents, and argumentative essays.

For all three subsets the annotators are university students, similar to the annotations from MultiLS \citet{shardlow2024readi}. We recruit a total of 90 native Romanian young adults with backgrounds in history, linguistics, and computer science. Using the Labelbox platform,\footnote{\url{https://labelbox.com}} we present each sentence and target word in a randomized trial. Annotators assign one of five categorical labels: \textit{very familiar}, \textit{simple}, \textit{neutral}, \textit{difficult}, or \textit{very difficult} to each target word. Before annotating the data, a trial of 15 samples is provided for practice and to explain the annotation guidelines. Following the Complex2.0 guidelines \cite{shardlow2022predicting}, we convert these labels to numerical values in the range [0, 1] and compute the average across annotators. Each word receives an average of 7.5 annotations, resulting in an overall inter-annotator agreement of Krippendorff’s $\alpha = 0.37$ (with quadratic weights). Since lexical complexity is subjective and the group of annotators is heterogeneous, we do not expect complete agreement. Each participant rates word complexity based on personal judgment, and the final score for each word is the mean of all individual ratings. \autoref{tab:stats} and \autoref{fig:densities} in the Appendix indicate a distribution shift in complexity following the translation of the English dataset into Romanian. Consistent with expectations and the findings from the MLSP dataset \cite{shardlow-etal-2024-extensible}, the English data exhibits a strong negative correlation ($-0.74$) between Zipf frequency \cite{wordfreq} and complexity. This confirms the general principle that more frequent words tend to be simpler. By contrast, the Romanian datasets deviate from this pattern, containing a larger proportion of words annotated as simple, with a moderately negative correlation ($-0.53 \pm .01$) with Zipf frequency.



\begin{figure}[htb]
    \centering
    \includegraphics[width=0.9\columnwidth]{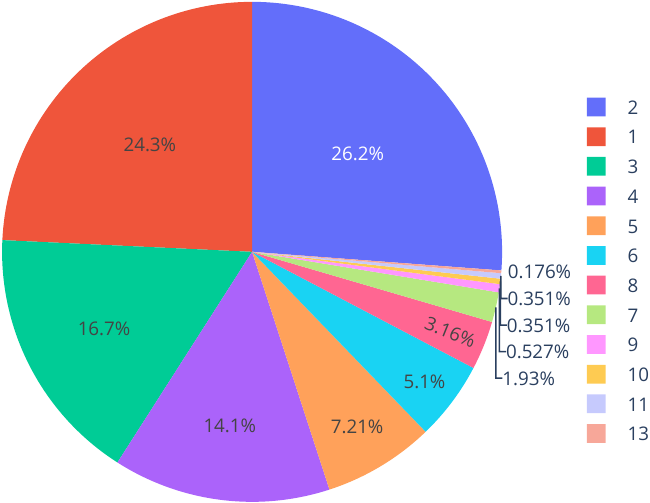} 
    \caption{Percentage of examples with $K$ lexical substitution candidates. Almost a quarter of sentences have no substitution candidates and almost half of sentences have at least 3 substitution candidates.}
    \label{fig:perc}
\end{figure}

\subsection{RoLS Dataset of Lexical Substitution Candidates}

In the original English guidelines \cite{shardlow2024readi}, each annotator provides one to three replacement candidates for a given word. These candidates are then ranked by the number of suggestions, with the most frequently suggested word treated as the simplest. This design choice impacts evaluation metrics such as MAP@K, which are sensitive to the ordering of the top-K candidates. In addition, suggestion frequency may not always be a reliable indicator of simplicity. Annotator creativity, tied frequencies (observed in approximately 30\% of the English data), and cases where less common words are actually simpler can all influence the results. 

In our approach, two new annotators from the same target group suggest replacement candidates for the HT data without judging their simplicity. They can use external resources, including dictionaries and LLMs. The final list is the union of both sets of suggestions, verified by a third annotator. About 25\% of target words have no suitable replacements, while nearly 50\% have at least three (see \autoref{fig:perc}). 

We then generate all possible sentence pairs from the suggested replacements and present them in random order to two additional annotators for pairwise simplicity judgments (Krippendorff’s $\alpha = 0.53$). Each annotator reviews approximately 3,000 pairs. Using the method from \citet{jerdee2023luck}, we apply a logistic Bradley–Terry model to compute sentence rankings based on pairwise comparisons (see \autoref{app:pair}). This ranking methodology is more robust than the one used for English, as it requires annotators to read and compare two complete sentences with candidate words replaced in context.

\begin{table}[!tb]
\centering
\resizebox{\columnwidth}{!}{%
\begin{tabular}{l|ccc}
\textbf{Train \textbackslash Test} & \textbf{HT} & \textbf{WT} & \textbf{RoLCP} \\
\hline
HT     & $0.56 \pm .11$ & $0.54$ & $0.52$ \\
WT     & $0.50$            & $0.68 \pm .04$ & $0.60$ \\
RoLCP  & $0.54$            & $0.61$ & $0.71 \pm .05$ \\
\end{tabular}
}
\caption{Performance of models trained on one dataset and tested on the others. 
Diagonal entries report 10-fold average and std. grouped 10-fold cross-validation results for each dataset (HT, WT, RoLCP). The model shows limited generalization across datasets.}
\label{tab:cv-results}
\end{table}

\begin{table*}[h]
\centering
\resizebox{.8\textwidth}{!}{%
\begin{tabular}{c|ccccc}
\textbf{Metric}    & \textbf{Apertus-8B} & \textbf{RoLlama-8B} & \textbf{DexFlex *} & \textbf{GPT-4o} & \textbf{BERT-Ro} \\ \hline
MAP@1              & 0.21                & 0.4                 & \textbf{0.41}      & 0.28            & 0.27             \\
MAP@3              & 0.11                & 0.18                & \textbf{0.31}      & 0.16            & 0.16             \\
MAP@5              & 0.10                & 0.16                & \textbf{0.3}       & 0.15            & 0.15             \\
MAP@10             & 0.10                & 0.16                & \textbf{0.33}      & 0.15            & 0.14             \\
Potential@3        & 0.29                & 0.49                & \textbf{0.6}       & 0.43            & 0.42             \\
Potential@5        & 0.33                & 0.51                & \textbf{0.67}      & 0.51            & 0.48             \\
Potential@10       & 0.39                & 0.52                & \textbf{0.7}       & 0.58            & 0.48             \\
ACC@1@top\_gold\_1 & 0.11                & \textbf{0.22}       & 0.15               & 0.16            & 0.14             \\
ACC@2@top\_gold\_1 & 0.15                & \textbf{0.26}       & 0.23               & 0.19            & 0.2              \\
ACC@3@top\_gold\_1 & 0.18                & 0.27                & \textbf{0.28}      & 0.23            & 0.26             \\ \hline
\end{tabular}
}
\caption{DexFlex consistently outperforms all other approaches across MAP and Potential metrics, showing strong robustness for synonym generation. While RoLlama achieves slightly better accuracy on top-gold metrics, it lags behind in overall ranking and coverage. The results suggest that a hybrid approach like DexFlex can be more effective than large general-purpose LLMs or fine-tuned BERT, especially when training on small datasets.
}
\label{tab:synonym_evaluation}
\end{table*}

\section{Results}
\paragraph{Lexical Complexity Prediction:} we employ a simple Ridge regressor baseline with handcrafted features: 1. zipf\_frequency from wordfreq library \cite{wordfreq}; 2. character length, number of vowels, approximate number of syllables from pyphen library;\footnote{\url{https://doc.courtbouillon.org/pyphen}}  3. the number of immediate children in syntactic dependency parse and the static embeddings from spaCy \texttt{ro\_core\_news\_lg} \cite{spacy}; 4. additional boolean features such as: is title, is entity, is sentence start, is sentence end regarded as markers of conceptual complexity \cite{stajner-etal-2020-coco}. 
We choose this approach for its simplicity and because it was one of the top-performing methods for Multilingual Lexical Complexity Prediction \cite{shardlow2024bea}.

We evaluate the models with the Pearson correlation coefficient, applying grouped 10-fold cross-validation so that no sentence in the training set appears in the test set, and we report cross-dataset evaluation scores in \autoref{tab:cv-results}. Cross-validation shows that RoLCP is the most consistent dataset ($r=0.71$), while HT is the most challenging ($r=0.56$). Off-diagonal results indicate that models generalize moderately well across datasets, with WT $\rightarrow$ RoLCP ($0.60$) and RoLCP $\rightarrow$ WT ($0.61$) showing the best cross-dataset transfer. The results are comparable to those reported for other languages by \citet{shardlow2024bea}: significantly lower than English ($0.85$), similar to Spanish ($0.72$) and German ($0.71$), close to French, Italian, and Catalan ($\simeq0.62$), and higher than Filipino ($0.56$) and Sinhala ($0.30$). For all these languages, the reported scores rely on deep learning methods.

\paragraph{Lexical Simplification:} the state-of-the-art approaches are strongly based on prompting external LLMs, as shown in the most recent Multilingual Lexical Simplification Pipeline (MLSP) Shared Task \cite{enomoto-etal-2024-tmu}.
We employ two open models: (1) Apertus-8B-Instruct-2509 from the Swiss AI Initiative \cite{apertus}, a massively multilingual model in which Romanian is represented through the FineWeb corpus \cite{penedo2025fineweb2} with 54 million tokens or 1.19\% of its training data; and (2) RoLlama3.1-8B from OpenLLM-Ro \cite{masala2024openllm}, a model based on the Llama3.1-8B instruction tuned on Romanian data. The prompt (see \autoref{app:prompt}) is written in English, as it yielded better results than its Romanian counterpart, and includes the full sentence together with the target word. The models are tasked with generating a JSON object containing ordered candidate replacements. For comparison, we apply the same strategy using GPT-4o to evaluate the performance gap between open models and closed-source systems.

In addition, we train a Romanian BERT model \cite{dumitrescu2020birth} with cross-entropy loss to make the model predict each replacement candidate. The model is evaluated with 5-fold grouped cross-validation, and training is performed for 3 epochs with a batch size of 16 and a learning rate of $0.0006$.


\paragraph{The DexFlex Framework} is a quasi-rule-based simplification system developed as an extension of the spaCy library. It automates tasks such as grammatical processing and synonym suggestion, using information from the open-source dexonline dictionary.\footnote{
\url{https://github.com/petruTH/DexFlex}
DEX is the acronym for the Explanatory Dictionary of Romanian, published and updated since 1975, available online in different variants at \url{https://dexonline.ro.}} DexFlex uses spaCy \cite{spacy} to identify the part-of-speech and grammatical features (gender, number, and person) of the target word and selects synonyms from the dictionary based on the similarity of contextual word embeddings from BERT \cite{dumitrescu2020birth}. The selected synonyms are properly inflected using dictionary information to be adequate in the context of the sentence (see \autoref{sec:dexapp}).
The LCP pipeline is used exclusively in conjunction with DexFlex to rank the candidate synonyms. 

\paragraph{Evaluation} is carried out using three metrics \cite{shardlow2024bea}: Mean Average Precision (MAP@N) evaluates a model's precision by assessing how well it ranks the correct class among the top N predictions. Potential@N measures the likelihood of finding relevant items within the top N results, and Accuracy@N@top\_gold\_1 assesses how often the first most likely substitution appears within the top N highly probable predictions. \autoref{tab:synonym_evaluation} contains the evaluation results across these metrics.
DexFlex consistently outperforms other approaches; however, the overall scores are considerably lower than what one might expect for a high-resource language (MAP@1 $\geq 0.72$). Evaluation scores are in the ranges of other low resource languages \cite{north2024deep} such as Sinhala ($\simeq 0.31$) and Filipino ($\simeq 0.36$). 


\section{Conclusions}
Our work introduces the first text simplification resources for Romanian and highlights key challenges in developing tools for under-represented languages. Our analyses show that cross-lingual transfer of complexity scores is not a viable resource creation procedure, causing distribution shifts. Furthermore, a hybrid rule-based system with synonym and inflection dictionaries offers a state-of-the-art solution for Romanian lexical simplification. This method is both more ecologically sustainable and linguistically grounded, while also outperforming prompt-based approaches with the latest large language models. Finally, since lexical complexity can be reliably predicted using hand-crafted features with performance comparable to LLM-based models \cite{shardlow2024bea}, we advocate for the development of simpler baseline models and for the integration of dictionaries into contemporary NLP pipelines wherever feasible.

\section{Limitations}
\label{sec:limitations}
The creation of the datasets was a long-term process during which we developed the annotation standards, and, as such, the three Romanian LCP datasets have several differences: (1) the initial human translation and word translation datasets are completely annotated by five annotators, while the RoLCP dataset is annotated by a pool of 80 annotators, each contributing to random subsets of the data, resulting in 10 annotations per word. For the RoLS dataset, due to the time-consuming nature of identifying candidate substitutions and providing human judgments for the large number of pairwise candidate comparisons, this process was only completed for the HT dataset.

Since the target annotators are educated young adults aged 20-33, the complexity signals captured in the dataset may limit the generalizability of models trained on this data for broader real-world applications. 


DexFlex has limitations, particularly in handling words with multiple parts of speech that share the same form. To address this, the framework uses spaCy to extract additional grammatical attributes in order for the correct part of speech and inflection to be applied to suggested synonyms. Certain nuances, such as distinguishing between similar articulate nouns like ``copacul'' (English: \textit{the tree}) and ``copacu''  remain challenging due to database inconsistencies. 


The total budget for running the experiments and conducting the data collection was 10\$.

\section{Ethics Statement}
The manual labeling was carried out by volunteers who agreed to annotate the data at no cost, and we are grateful for their significant contribution. Participants were invited via email and some students used the collected data to develop their dissertations or to build in-class projects. The annotators agreed to publish labels along with the dataset under anonymity. 

The texts we used for creating the dataset were sourced from platforms like Wikipedia, Wikibooks, and other public online sources. These sources either reside in the public domain or are published under permissive licenses (such as Creative Commons) or allow for academic fair use, i.e., small excerpts for research and the creation of derivative works.

We release our data and code under the CC BY-NC-SA 4.0 license.


\section*{Acknowledgments}
We express our gratitude to the annotators whose labor was fundamental in building the Romanian dataset.
We thank Oleksandra Kuvshynova, Mircea Marin, Radu Ciobanu, Anamaria Hodivoianu, and Ana Sabina Uban for their valuable support during the process of writing and refining this work. 
We would also like to thank the anonymous reviewers and the area chair for their constructive feedback.

This research is mainly supported by InstRead: Research Instruments for the Text Complexity, Simplification and Readability Assessment  CNCS - UEFISCDI project number PN-IV-P2-2.1-TE-2023-2007 and by the project ``Romanian Hub for Artificial Intelligence - HRIA'', Smart Growth, Digitization and Financial Instruments Program, 2021-2027, MySMIS no. 334906.

\bibliography{custom,literacy_motivation}

@inproceedings{specia2012semeval,
  title={Semeval-2012 task 1: English lexical simplification},
  author={Specia, Lucia and Jauhar, Sujay Kumar and Mihalcea, Rada},
  booktitle={* SEM 2012: The First Joint Conference on Lexical and Computational Semantics--Volume 1: Proceedings of the main conference and the shared task, and Volume 2: Proceedings of the Sixth International Workshop on Semantic Evaluation (SemEval 2012)},
  pages={347--355},
  year={2012}
}

@inproceedings{paetzold-specia-2016-semeval,
    title = "{S}em{E}val 2016 Task 11: Complex Word Identification",
    author = "Paetzold, Gustavo  and
      Specia, Lucia",
    editor = "Bethard, Steven  and
      Carpuat, Marine  and
      Cer, Daniel  and
      Jurgens, David  and
      Nakov, Preslav  and
      Zesch, Torsten",
    booktitle = "Proceedings of the 10th International Workshop on Semantic Evaluation ({S}em{E}val-2016)",
    month = jun,
    year = "2016",
    address = "San Diego, California",
    publisher = "Association for Computational Linguistics",
    url = "https://aclanthology.org/S16-1085",
    doi = "10.18653/v1/S16-1085",
    pages = "560--569",
}

@inproceedings{yimam-etal-2018-report,
    title = "A Report on the Complex Word Identification Shared Task 2018",
    author = {Yimam, Seid Muhie  and
      Biemann, Chris  and
      Malmasi, Shervin  and
      Paetzold, Gustavo  and
      Specia, Lucia  and
      {\v{S}}tajner, Sanja  and
      Tack, Ana{\"\i}s  and
      Zampieri, Marcos},
    editor = "Tetreault, Joel  and
      Burstein, Jill  and
      Kochmar, Ekaterina  and
      Leacock, Claudia  and
      Yannakoudakis, Helen",
    booktitle = "Proceedings of the Thirteenth Workshop on Innovative Use of {NLP} for Building Educational Applications",
    month = jun,
    year = "2018",
    address = "New Orleans, Louisiana",
    publisher = "Association for Computational Linguistics",
    url = "https://aclanthology.org/W18-0507",
    doi = "10.18653/v1/W18-0507",
    pages = "66--78",
}

@inproceedings{saggion2023findings,
    title = "Findings of the {TSAR}-2022 Shared Task on Multilingual Lexical Simplification",
    author = "Saggion, Horacio  and
      {\v{S}}tajner, Sanja  and
      Ferr{\'e}s, Daniel  and
      Sheang, Kim Cheng  and
      Shardlow, Matthew  and
      North, Kai  and
      Zampieri, Marcos",
    editor = "{\v{S}}tajner, Sanja  and
      Saggion, Horacio  and
      Ferr{\'e}s, Daniel  and
      Shardlow, Matthew  and
      Sheang, Kim Cheng  and
      North, Kai  and
      Zampieri, Marcos  and
      Xu, Wei",
    booktitle = "Proceedings of the Workshop on Text Simplification, Accessibility, and Readability (TSAR-2022)",
    month = dec,
    year = "2022",
    address = "Abu Dhabi, United Arab Emirates (Virtual)",
    publisher = "Association for Computational Linguistics",
    url = "https://aclanthology.org/2022.tsar-1.31",
    doi = "10.18653/v1/2022.tsar-1.31",
    pages = "271--283",
}

@proceedings{tsar-2024-text,
    title = "Proceedings of the Third Workshop on Text Simplification, Accessibility and Readability",
    editor = "{\v{S}}tajner, Sanja  and
      Saggio, Horacio  and
      Shardlow, Matthew  and
      Alva-Manchego, Fernando",
    month = nov,
    year = "2024",
    address = "Miami, Florida",
    publisher = "EMNLP 2024",
}

@inproceedings{shardlow2021semeval,
    title = "{S}em{E}val-2021 Task 1: Lexical Complexity Prediction",
    author = "Shardlow, Matthew  and
      Evans, Richard  and
      Paetzold, Gustavo Henrique  and
      Zampieri, Marcos",
    editor = "Palmer, Alexis  and
      Schneider, Nathan  and
      Schluter, Natalie  and
      Emerson, Guy  and
      Herbelot, Aurelie  and
      Zhu, Xiaodan",
    booktitle = "Proceedings of the 15th International Workshop on Semantic Evaluation (SemEval-2021)",
    month = aug,
    year = "2021",
    address = "Online",
    publisher = "Association for Computational Linguistics",
    url = "https://aclanthology.org/2021.semeval-1.1/",
    doi = "10.18653/v1/2021.semeval-1.1",
    pages = "1--16"
}

@inproceedings{shardlow2024bea,
title={{The BEA 2024 Shared Task on the Multilingual Lexical Simplification Pipeline}},
author={Shardlow, Matthew and Alva-Manchego, Fernando and Batista-Navarro, Riza and Bott, Stefan and Calderon Ramirez, Saul and Cardon, Rémi and François, Thomas and Hayakawa, Akio and Horbach, Andrea and Huelsing, Anna and Ide, Yusuke and Imperial, Joseph Marvin and Nohejl, Adam and North, Kai and Occhipinti, Laura and Peréz Rojas, Nelson and Raihan, Nishat and Ranasinghe, Tharindu and Solis Salazar, Martin and \v{S}tajner, Sanja and Zampieri, Marcos and Saggion, Horacio},
booktitle={Proceedings of the 19th Workshop on Innovative Use of NLP for Building Educational Applications (BEA)},
year={2024}
}

@inproceedings{shardlow-etal-2024-extensible,
    title = "An Extensible Massively Multilingual Lexical Simplification Pipeline Dataset using the {M}ulti{LS} Framework",
    author = {Shardlow, Matthew  and
      Alva-Manchego, Fernando  and
      Batista-Navarro, Riza  and
      Bott, Stefan  and
      Calderon Ramirez, Saul  and
      Cardon, R{\'e}mi  and
      Fran{\c{c}}ois, Thomas  and
      Hayakawa, Akio  and
      Horbach, Andrea  and
      H{\"u}lsing, Anna  and
      Ide, Yusuke  and
      Imperial, Joseph Marvin  and
      Nohejl, Adam  and
      North, Kai  and
      Occhipinti, Laura  and
      Per{\'e}z Rojas, Nelson  and
      Raihan, Nishat  and
      Ranasinghe, Tharindu  and
      Solis Salazar, Martin  and
      Zampieri, Marcos  and
      Saggion, Horacio},
    editor = "Wilkens, Rodrigo  and
      Cardon, R{\'e}mi  and
      Todirascu, Amalia  and
      Gala, N{\'u}ria",
    booktitle = "Proceedings of the 3rd Workshop on Tools and Resources for People with REAding DIfficulties (READI) @ LREC-COLING 2024",
    month = may,
    year = "2024",
    address = "Torino, Italia",
    publisher = "ELRA and ICCL",
    url = "https://aclanthology.org/2024.readi-1.4",
    pages = "38--46",
}

@inproceedings{specia2010translating,
  title={Translating from complex to simplified sentences},
  author={Specia, Lucia},
  booktitle={Computational Processing of the Portuguese Language: 9th International Conference, PROPOR 2010, Porto Alegre, RS, Brazil, April 27-30, 2010. Proceedings 9},
  pages={30--39},
  year={2010},
  organization={Springer}
}

@inproceedings{nisioi-etal-2017-exploring,
    title = "Exploring Neural Text Simplification Models",
    author = "Nisioi, Sergiu  and
      {\v{S}}tajner, Sanja  and
      Ponzetto, Simone Paolo  and
      Dinu, Liviu P.",
    editor = "Barzilay, Regina  and
      Kan, Min-Yen",
    booktitle = "Proceedings of the 55th Annual Meeting of the Association for Computational Linguistics (Volume 2: Short Papers)",
    month = jul,
    year = "2017",
    address = "Vancouver, Canada",
    publisher = "Association for Computational Linguistics",
    url = "https://aclanthology.org/P17-2014",
    doi = "10.18653/v1/P17-2014",
    pages = "85--91",
}

@inproceedings{stajner-etal-2023-less,
    title = "{L}e{SS}: A Computationally-Light Lexical Simplifier for {S}panish",
    author = "Stajner, Sanja  and
      Ibanez, Daniel  and
      Saggion, Horacio",
    editor = "Mitkov, Ruslan  and
      Angelova, Galia",
    booktitle = "Proceedings of the 14th International Conference on Recent Advances in Natural Language Processing",
    month = sep,
    year = "2023",
    address = "Varna, Bulgaria",
    publisher = "INCOMA Ltd., Shoumen, Bulgaria",
    url = "https://aclanthology.org/2023.ranlp-1.120",
    pages = "1132--1142",
}

@inproceedings{dou2023automatic,
  title={Automatic and Human-AI Interactive Text Generation (with a focus on Text Simplification and Revision)},
  author={Dou, Yao and Laban, Philippe and Gardent, Claire and Xu, Wei},
  booktitle={Proceedings of the 62nd Annual Meeting of the Association for Computational Linguistics (Volume 5: Tutorial Abstracts)},
  pages={3--4},
  year={2024}
}

@inproceedings{gooding2022ethical,
    title = "On the Ethical Considerations of Text Simplification",
    author = "Gooding, Sian",
    editor = "Ebling, Sarah  and
      Prud{'}hommeaux, Emily  and
      Vaidyanathan, Preethi",
    booktitle = "Ninth Workshop on Speech and Language Processing for Assistive Technologies (SLPAT-2022)",
    month = may,
    year = "2022",
    address = "Dublin, Ireland",
    publisher = "Association for Computational Linguistics",
    url = "https://aclanthology.org/2022.slpat-1.7/",
    doi = "10.18653/v1/2022.slpat-1.7",
    pages = "50--57"
}

@inproceedings{carroll1998practical,
  title={Practical simplification of English newspaper text to assist aphasic readers},
  author={Carroll, John and Minnen, Guido and Canning, Yvonne and Devlin, Siobhan and Tait, John},
  booktitle={Proceedings of the AAAI-98 workshop on integrating artificial intelligence and assistive technology},
  pages={7--10},
  year={1998},
  organization={Madison, WI}
}

@inproceedings{de2010text,
  title={Text simplification for children},
  author={De Belder, Jan and Moens, Marie-Francine},
  booktitle={Proceedings of the SIGIR workshop on accessible search systems},
  pages={19--26},
  year={2010}
}

@inproceedings{glavas-stajner-2015-simplifying,
    title = "Simplifying Lexical Simplification: Do We Need Simplified Corpora?",
    author = "Glava{\v{s}}, Goran  and
      {\v{S}}tajner, Sanja",
    editor = "Zong, Chengqing  and
      Strube, Michael",
    booktitle = "Proceedings of the 53rd Annual Meeting of the Association for Computational Linguistics and the 7th International Joint Conference on Natural Language Processing (Volume 2: Short Papers)",
    month = jul,
    year = "2015",
    address = "Beijing, China",
    publisher = "Association for Computational Linguistics",
    url = "https://aclanthology.org/P15-2011",
    doi = "10.3115/v1/P15-2011",
    pages = "63--68",
}

@inproceedings{lee-yeung-2018-personalizing,
    title = "Personalizing Lexical Simplification",
    author = "Lee, John  and
      Yeung, Chak Yan",
    editor = "Bender, Emily M.  and
      Derczynski, Leon  and
      Isabelle, Pierre",
    booktitle = "Proceedings of the 27th International Conference on Computational Linguistics",
    month = aug,
    year = "2018",
    address = "Santa Fe, New Mexico, USA",
    publisher = "Association for Computational Linguistics",
    url = "https://aclanthology.org/C18-1019",
    pages = "224--232",
}

@inproceedings{gooding-tragut-2022-one,
    title = "One Size Does Not Fit All: The Case for Personalised Word Complexity Models",
    author = "Gooding, Sian  and
      Tragut, Manuel",
    editor = "Carpuat, Marine  and
      de Marneffe, Marie-Catherine  and
      Meza Ruiz, Ivan Vladimir",
    booktitle = "Findings of the Association for Computational Linguistics: NAACL 2022",
    month = jul,
    year = "2022",
    address = "Seattle, United States",
    publisher = "Association for Computational Linguistics",
    url = "https://aclanthology.org/2022.findings-naacl.27",
    doi = "10.18653/v1/2022.findings-naacl.27",
    pages = "353--365",
}

@inproceedings{sheang-etal-2022-controllable,
    title = "Controllable Lexical Simplification for {E}nglish",
    author = "Sheang, Kim Cheng  and
      Ferr{\'e}s, Daniel  and
      Saggion, Horacio",
    editor = "{\v{S}}tajner, Sanja  and
      Saggion, Horacio  and
      Ferr{\'e}s, Daniel  and
      Shardlow, Matthew  and
      Sheang, Kim Cheng  and
      North, Kai  and
      Zampieri, Marcos  and
      Xu, Wei",
    booktitle = "Proceedings of the Workshop on Text Simplification, Accessibility, and Readability (TSAR-2022)",
    month = dec,
    year = "2022",
    address = "Abu Dhabi, United Arab Emirates (Virtual)",
    publisher = "Association for Computational Linguistics",
    url = "https://aclanthology.org/2022.tsar-1.19",
    doi = "10.18653/v1/2022.tsar-1.19",
    pages = "199--206",
}

@article{shardlow2022predicting,
  title={Predicting lexical complexity in English texts: the Complex 2.0 dataset},
  author={Shardlow, Matthew and Evans, Richard and Zampieri, Marcos},
  journal={Language Resources and Evaluation},
  volume={56},
  number={4},
  pages={1153--1194},
  year={2022},
  publisher={Springer}
}

@article{north2023lexical,
  title={Lexical complexity prediction: An overview},
  author={North, Kai and Zampieri, Marcos and Shardlow, Matthew},
  journal={ACM Computing Surveys},
  volume={55},
  number={9},
  pages={1--42},
  year={2023},
  publisher={ACM New York, NY}
}

@article{north2024deep,
  title={Deep learning approaches to lexical simplification: A survey},
  author={North, Kai and Ranasinghe, Tharindu and Shardlow, Matthew and Zampieri, Marcos},
  journal={Journal of Intelligent Information Systems},
  pages={1--24},
  year={2024},
  publisher={Springer}
}

@inproceedings{pintard-francois-2020-combining,
    title = "Combining Expert Knowledge with Frequency Information to Infer {CEFR} Levels for Words",
    author = "Pintard, Alice  and
      Fran{\c{c}}ois, Thomas",
    editor = "Gala, N{\'u}ria  and
      Wilkens, Rodrigo",
    booktitle = "Proceedings of the 1st Workshop on Tools and Resources to Empower People with REAding DIfficulties (READI)",
    month = may,
    year = "2020",
    address = "Marseille, France",
    publisher = "European Language Resources Association",
    url = "https://aclanthology.org/2020.readi-1.13",
    pages = "85--92",
    language = "English",
    ISBN = "979-10-95546-45-0",
}

@article{ebling2022automatic,
  title={Automatic text simplification for German},
  author={Ebling, Sarah and Battisti, Alessia and Kostrzewa, Marek and Pf{\"u}tze, Dominik and Rios, Annette and S{\"a}uberli, Andreas and Spring, Nicolas},
  journal={Frontiers in Communication},
  volume={7},
  pages={706718},
  year={2022},
  publisher={Frontiers Media SA}
}

@inproceedings{kodaira-etal-2016-controlled,
    title = "Controlled and Balanced Dataset for {J}apanese Lexical Simplification",
    author = "Kodaira, Tomonori  and
      Kajiwara, Tomoyuki  and
      Komachi, Mamoru",
    booktitle = "Proceedings of the {ACL} 2016 Student Research Workshop",
    year = "2016",
    pages = "1--7",
}

@inproceedings{ide2023,
  title     = "Japanese Lexical Complexity for Non-Native Readers: A New Dataset",
  booktitle = "Proceedings of the Eighteenth Workshop on Innovative Use of {NLP} for Building Educational Applications",
  author    = "Ide, Yusuke and Mita, Masato and Nohejl, Adam and Ouchi, Hiroki and Watanabe, Taro",
  month     = July,
  year      = 2023,
  publisher = "Association for Computational Linguistics",
}

@article{hartmann2020adaptaccao,
  title={Adapta{\c{c}}{\~a}o lexical autom{\'a}tica em textos informativos do portugu{\^e}s brasileiro para o ensino fundamental},
  author={Hartmann, Nathan Siegle and Alu{\'\i}sio, Sandra Maria},
  journal={Linguam{\'a}tica},
  volume={12},
  number={2},
  pages={3--27},
  year={2020}
}

@article{qiang2021chinese,
  title={Chinese lexical simplification},
  author={Qiang, Jipeng and Lu, Xinyu and Li, Yun and Yuan, Yunhao and Wu, Xindong},
  journal={IEEE/ACM Transactions on Audio, Speech, and Language Processing},
  volume={29},
  pages={1819--1828},
  year={2021},
  publisher={IEEE}
}

@inproceedings{ferres-saggion-2022-alexsis,
    title = "{ALEXSIS}: A Dataset for Lexical Simplification in {S}panish",
    author = "Ferr{\'e}s, Daniel  and
      Saggion, Horacio",
    editor = "Calzolari, Nicoletta  and
      B{\'e}chet, Fr{\'e}d{\'e}ric  and
      Blache, Philippe  and
      Choukri, Khalid  and
      Cieri, Christopher  and
      Declerck, Thierry  and
      Goggi, Sara  and
      Isahara, Hitoshi  and
      Maegaard, Bente  and
      Mariani, Joseph  and
      Mazo, H{\'e}l{\`e}ne  and
      Odijk, Jan  and
      Piperidis, Stelios",
    booktitle = "Proceedings of the Thirteenth Language Resources and Evaluation Conference",
    month = jun,
    year = "2022",
    address = "Marseille, France",
    publisher = "European Language Resources Association",
    url = "https://aclanthology.org/2022.lrec-1.383",
    pages = "3582--3594",
}

@article{alarcon2023easier,
  title={EASIER corpus: A lexical simplification resource for people with cognitive impairments},
  author={Alarcon, Rodrigo and Moreno, Lourdes and Mart{\'\i}nez, Paloma},
  journal={Plos one},
  volume={18},
  number={4},
  pages={e0283622},
  year={2023},
  publisher={Public Library of Science San Francisco, CA USA}
}

@inproceedings{billami2018resyf,
  title={ReSyf: a French lexicon with ranked synonyms},
  author={Billami, Mokhtar Boumedyen and Fran{\c{c}}ois, Thomas and Gala, N{\'u}ria},
  booktitle={27th International Conference on Computational Linguistics (COLING 2018)},
  year={2018}
}

@inproceedings{kajiwara2015evaluation,
  title={Evaluation dataset and system for Japanese lexical simplification},
  author={Kajiwara, Tomoyuki and Yamamoto, Kazuhide},
  booktitle={Proceedings of the ACL-IJCNLP 2015 Student Research Workshop},
  pages={35--40},
  year={2015}
}

@article{blum1978universals,
  title={Universals of lexical simplification},
  author={Blum, Shoshana and Levenston, Eddie A},
  journal={Language learning},
  volume={28},
  number={2},
  pages={399--415},
  year={1978},
  publisher={Wiley Online Library}
}

@inproceedings{rabinovich2016similarities,
  title={On the Similarities Between Native, Non-native and Translated Texts},
  author={Rabinovich, Ella and Nisioi, Sergiu and Ordan, Noam and Wintner, Shuly},
  booktitle={Proceedings of the 54th Annual Meeting of the Association for Computational Linguistics (Volume 1: Long Papers)},
  pages={1870--1881},
  year={2016}
}

@article{yao2024survey,
  title={A survey on large language model (llm) security and privacy: The good, the bad, and the ugly},
  author={Yao, Yifan and Duan, Jinhao and Xu, Kaidi and Cai, Yuanfang and Sun, Zhibo and Zhang, Yue},
  journal={High-Confidence Computing},
  pages={100211},
  year={2024},
  publisher={Elsevier}
}

@misc{AllenZhu-icml2024-tutorial,
    author = {{Allen-Zhu}, Zeyuan},
    title = {{ICML 2024 Tutorial: Physics of Language Models}},
    year = {2024},
    month = {July},
    note = {Project page: \url{https://physics.allen-zhu.com/}}
}

@inproceedings{codruț2024rodia,
  title={RoDia: A New Dataset for Romanian Dialect Identification from Speech},
  author={Codruț, Rotaru and Ristea, Nicolae and Ionescu, Radu},
  booktitle={Findings of the Association for Computational Linguistics: NAACL 2024},
  pages={279--286},
  year={2024}
}

@inproceedings{sastre-etal-2024-retuyt,
    title = "{RETUYT}-{INCO} at {MLSP} 2024: Experiments on Language Simplification using Embeddings, Classifiers and Large Language Models",
    author = "Sastre, Ignacio  and
      Alfonso, Leandro  and
      Fleitas, Facundo  and
      Gil, Federico  and
      Lucas, Andr{\'e}s  and
      Spoturno, Tom{\'a}s  and
      G{\'o}ngora, Santiago  and
      Ros{\'a}, Aiala  and
      Chiruzzo, Luis",
    editor = {Kochmar, Ekaterina  and
      Bexte, Marie  and
      Burstein, Jill  and
      Horbach, Andrea  and
      Laarmann-Quante, Ronja  and
      Tack, Ana{\"\i}s  and
      Yaneva, Victoria  and
      Yuan, Zheng},
    booktitle = "Proceedings of the 19th Workshop on Innovative Use of NLP for Building Educational Applications (BEA 2024)",
    month = jun,
    year = "2024",
    address = "Mexico City, Mexico",
    publisher = "Association for Computational Linguistics",
    url = "https://aclanthology.org/2024.bea-1.56",
    pages = "618--626",
}

@inproceedings{enomoto-etal-2024-tmu,
    title = "{TMU}-{HIT} at {MLSP} 2024: How Well Can {GPT}-4 Tackle Multilingual Lexical Simplification?",
    author = "Enomoto, Taisei  and
      Kim, Hwichan  and
      Hirasawa, Tosho  and
      Nagai, Yoshinari  and
      Sato, Ayako  and
      Nakajima, Kyotaro  and
      Komachi, Mamoru",
    editor = {Kochmar, Ekaterina  and
      Bexte, Marie  and
      Burstein, Jill  and
      Horbach, Andrea  and
      Laarmann-Quante, Ronja  and
      Tack, Ana{\"\i}s  and
      Yaneva, Victoria  and
      Yuan, Zheng},
    booktitle = "Proceedings of the 19th Workshop on Innovative Use of NLP for Building Educational Applications (BEA 2024)",
    month = jun,
    year = "2024",
    address = "Mexico City, Mexico",
    publisher = "Association for Computational Linguistics",
    url = "https://aclanthology.org/2024.bea-1.52",
    pages = "590--598",
}

@inproceedings{cristeaNisioi24,
    title = "Archaeology at MLSP 2024: Machine Translation for Lexical Complexity Prediction and Lexical Simplification",
    author = "Cristea, Petru  and
      Nisioi, Sergiu",
    editor = {Kochmar, Ekaterina  and
      Bexte, Marie  and
      Burstein, Jill  and
      Horbach, Andrea  and
      Laarmann-Quante, Ronja  and
      Tack, Ana{\"\i}s  and
      Yaneva, Victoria  and
      Yuan, Zheng},
    booktitle = "Proceedings of the 19th Workshop on Innovative Use of NLP for Building Educational Applications (BEA 2024)",
    month = jun,
    year = "2024",
    address = "Mexico City, Mexico",
    publisher = "Association for Computational Linguistics",
    url = "https://aclanthology.org/2024.bea-1.55",
    pages = "610--617",
}

@inproceedings{zilio-etal-2020-lexical,
    title = "A Lexical Simplification Tool for Promoting Health Literacy",
    author = "Zilio, Leonardo  and
      Braga Paraguassu, Liana  and
      Leiva Hercules, Luis Antonio  and
      Ponomarenko, Gabriel  and
      Berwanger, Laura  and
      Bocorny Finatto, Maria Jos{\'e}",
    editor = "Gala, N{\'u}ria  and
      Wilkens, Rodrigo",
    booktitle = "Proceedings of the 1st Workshop on Tools and Resources to Empower People with REAding DIfficulties (READI)",
    month = may,
    year = "2020",
    address = "Marseille, France",
    publisher = "European Language Resources Association",
    url = "https://aclanthology.org/2020.readi-1.11",
    pages = "70--76",
    language = "English",
    ISBN = "979-10-95546-45-0",
}

@article{vstajner2021automatic,
  title={Automatic text simplification for social good: Progress and challenges},
  author={{\v{S}}tajner, Sanja},
  journal={Findings of the Association for Computational Linguistics: ACL-IJCNLP 2021},
  pages={2637--2652},
  year={2021}
}

@inproceedings{shardlow2024readi,
title={{An Extensible Massively Multilingual Lexical Simplification Pipeline Dataset using the MultiLS Framework}},
author={Shardlow, Matthew and Alva-Manchego, Fernando and Batista-Navarro, Riza and Bott, Stefan and Calderon Ramirez, Saul and Cardon, Rémi and François, Thomas and Hayakawa, Akio and Horbach, Andrea and Huelsing, Anna and Ide, Yusuke and Imperial, Joseph Marvin and Nohejl, Adam and North, Kai and Occhipinti, Laura and Peréz Rojas, Nelson and Raihan, Nishat and Ranasinghe, Tharindu and Solis Salazar, Martin and Zampieri, Marcos and Saggion, Horacio},
booktitle={Proceedings of the 3rd Workshop on Tools and Resources for People with REAding DIfficulties (READI)},
year={2024}
}

@software{spacy,
  author       = {Ines Montani and
                  Matthew Honnibal and
                  Matthew Honnibal and
                  Adriane Boyd and
                  Sofie Van Landeghem and
                  Henning Peters},
  title        = {{explosion/spaCy: v3.7.2: Fixes for APIs and 
                   requirements}},
  month        = oct,
  year         = 2023,
  publisher    = {Zenodo},
  version      = {v3.7.2},
  doi          = {10.5281/zenodo.10009823},
  url          = {https://doi.org/10.5281/zenodo.10009823}
}

@software{wordfreq,
  author       = {Robyn Speer},
  title        = {rspeer/wordfreq: v3.0},
  month        = sep,
  year         = 2022,
  publisher    = {Zenodo},
  version      = {v3.0.2},
  doi          = {10.5281/zenodo.7199437},
  url          = {https://doi.org/10.5281/zenodo.7199437}
}

@article{devlin1998use,
  title={The use of a psycholinguistic database in the simplification of text for aphasic readers},
  author={Devlin, Siobhan},
  journal={Linguistic databases},
  year={1998},
  publisher={CSLI}
}

@inproceedings{stajner-etal-2020-coco,
    title = "{C}o{C}o: A Tool for Automatically Assessing Conceptual Complexity of Texts",
    author = "Stajner, Sanja  and
      Nisioi, Sergiu  and
      Hulpu{\textcommabelow{s}}, Ioana",
    editor = "Calzolari, Nicoletta  and
      B{\'e}chet, Fr{\'e}d{\'e}ric  and
      Blache, Philippe  and
      Choukri, Khalid  and
      Cieri, Christopher  and
      Declerck, Thierry  and
      Goggi, Sara  and
      Isahara, Hitoshi  and
      Maegaard, Bente  and
      Mariani, Joseph  and
      Mazo, H{\'e}l{\`e}ne  and
      Moreno, Asuncion  and
      Odijk, Jan  and
      Piperidis, Stelios",
    booktitle = "Proceedings of the Twelfth Language Resources and Evaluation Conference",
    month = may,
    year = "2020",
    address = "Marseille, France",
    publisher = "European Language Resources Association",
    url = "https://aclanthology.org/2020.lrec-1.887",
    pages = "7179--7186",
    language = "English",
    ISBN = "979-10-95546-34-4",
}

@article{masala2024openllm,
  title={“Vorbești Rom{\^a}nește?” A Recipe to Train Powerful Romanian LLMs with English Instructions},
  author={Masala, Mihai and Ilie-Ablachim, Denis and Dima, Alexandru and Corlatescu, Dragos Georgian and Zavelca, Miruna-Andreea and Olaru, Ovio and Terian, Simina-Maria and Terian, Andrei and Leordeanu, Marius and Velicu, Horia and others},
  booktitle={Findings of the Association for Computational Linguistics: EMNLP 2024},
  pages={11632--11647},
  year={2024}
}

@inproceedings{hobo-etal-2023-geen,
    title = "{``}Geen makkie{''}: Interpretable Classification and Simplification of {D}utch Text Complexity",
    author = "Hobo, Eliza  and
      Pouw, Charlotte  and
      Beinborn, Lisa",
    editor = {Kochmar, Ekaterina  and
      Burstein, Jill  and
      Horbach, Andrea  and
      Laarmann-Quante, Ronja  and
      Madnani, Nitin  and
      Tack, Ana{\"\i}s  and
      Yaneva, Victoria  and
      Yuan, Zheng  and
      Zesch, Torsten},
    booktitle = "Proceedings of the 18th Workshop on Innovative Use of NLP for Building Educational Applications (BEA 2023)",
    month = jul,
    year = "2023",
    address = "Toronto, Canada",
    publisher = "Association for Computational Linguistics",
    url = "https://aclanthology.org/2023.bea-1.42",
    doi = "10.18653/v1/2023.bea-1.42",
    pages = "503--517",
}

@inproceedings{dumitrescu2020birth,
  title={The birth of Romanian BERT},
  author={Dumitrescu, Stefan and Avram, Andrei-Marius and Pyysalo, Sampo},
  booktitle={Findings of the Association for Computational Linguistics: EMNLP 2020},
  pages={4324--4328},
  year={2020}
}

@article{ferres2017adaptable,
  title={An Adaptable Lexical Simplification Architecture for Major Ibero-Romance Languages},
  author={Ferr{\'e}s, Daniel and Saggion, Horacio and Guinovart, Xavier G{\'o}mez},
  journal={EMNLP 2017},
  pages={40},
  year={2017}
}

@inproceedings{qiang2020lexical,
  title={Lexical simplification with pretrained encoders},
  author={Qiang, Jipeng and Li, Yun and Zhu, Yi and Yuan, Yunhao and Wu, Xindong},
  booktitle={Proceedings of the AAAI Conference on Artificial Intelligence},
  volume={34},
  number={05},
  pages={8649--8656},
  year={2020}
}

@article{qiang2021lsbert,
  title={Lsbert: Lexical simplification based on bert},
  author={Qiang, Jipeng and Li, Yun and Zhu, Yi and Yuan, Yunhao and Shi, Yang and Wu, Xindong},
  journal={IEEE/ACM transactions on audio, speech, and language processing},
  volume={29},
  pages={3064--3076},
  year={2021},
  publisher={IEEE}
}

@inproceedings{baez-saggion-2023-lsllama,
    title = "{LSL}lama: Fine-Tuned {LL}a{MA} for Lexical Simplification",
    author = "Baez, Anthony  and
      Saggion, Horacio",
    editor = "{\v{S}}tajner, Sanja  and
      Saggio, Horacio  and
      Shardlow, Matthew  and
      Alva-Manchego, Fernando",
    booktitle = "Proceedings of the Second Workshop on Text Simplification, Accessibility and Readability",
    month = sep,
    year = "2023",
    address = "Varna, Bulgaria",
    publisher = "INCOMA Ltd., Shoumen, Bulgaria",
    url = "https://aclanthology.org/2023.tsar-1.10",
    pages = "102--108",
}

@article{sheang2023multilingual,
  title={Multilingual Controllable Transformer-Based Lexical Simplification},
  author={Sheang, Kim Cheng and Saggion, Horacio},
  journal={Procesamiento del Lenguaje Natural},
  volume={71},
  pages={109},
  year={2023},
  publisher={Sociedad Espa{\~n}ola para el Procesamiento del Lenguaje Natural}
}

@inproceedings{paetzold-specia-2015-lexenstein,
    title = "{LEX}enstein: A Framework for Lexical Simplification",
    author = "Paetzold, Gustavo  and
      Specia, Lucia",
    editor = "Chen, Hsin-Hsi  and
      Markert, Katja",
    booktitle = "Proceedings of {ACL}-{IJCNLP} 2015 System Demonstrations",
    month = jul,
    year = "2015",
    address = "Beijing, China",
    publisher = "Association for Computational Linguistics and The Asian Federation of Natural Language Processing",
    url = "https://aclanthology.org/P15-4015",
    doi = "10.3115/v1/P15-4015",
    pages = "85--90",
}

@inproceedings{north2022alexsis,
  title={ALEXSIS-PT: A New Resource for Portuguese Lexical Simplification},
  author={North, Kai and Zampieri, Marcos and Ranasinghe, Tharindu},
  booktitle={Proceedings-International Conference on Computational Linguistics, COLING},
  volume={29},
  number={1},
  pages={6057--6062},
  year={2022}
}

@inproceedings{aumiller-gertz-2022-unihd,
    title = "{U}ni{HD} at {TSAR}-2022 Shared Task: Is Compute All We Need for Lexical Simplification?",
    author = "Aumiller, Dennis  and
      Gertz, Michael",
    editor = "{\v{S}}tajner, Sanja  and
      Saggion, Horacio  and
      Ferr{\'e}s, Daniel  and
      Shardlow, Matthew  and
      Sheang, Kim Cheng  and
      North, Kai  and
      Zampieri, Marcos  and
      Xu, Wei",
    booktitle = "Proceedings of the Workshop on Text Simplification, Accessibility, and Readability (TSAR-2022)",
    month = dec,
    year = "2022",
    address = "Abu Dhabi, United Arab Emirates (Virtual)",
    publisher = "Association for Computational Linguistics",
    url = "https://aclanthology.org/2022.tsar-1.28",
    doi = "10.18653/v1/2022.tsar-1.28",
    pages = "251--258",
}

@article{jerdee2023luck,
  author       = {Jerdee, M. and Newman, M. E. J.},
  title        = {Luck, skill, and depth of competition in games and social hierarchies},
  journal      = {Science Advances},
  year         = {2024},
  volume       = {10},
  number       = {45},
  pages        = {eadn2654},
  doi          = {10.1126/sciadv.adn2654},
  url          = {https://doi.org/10.1126/sciadv.adn2654},
  note         = {Epub 2024 Nov 6. PMID: 39504380; PMCID: PMC11540035}
}

@inproceedings{midrigan-ciochina-etal-2020-resources,
    title = "Resources in Underrepresented Languages: Building a Representative {R}omanian Corpus",
    author = "Midrigan - Ciochina, Ludmila  and
      Boyd, Victoria  and
      Sanchez-Ortega, Lucila  and
      Malancea{\_}Malac, Diana  and
      Midrigan, Doina  and
      Corina, David P.",
    editor = "Calzolari, Nicoletta  and
      B{\'e}chet, Fr{\'e}d{\'e}ric  and
      Blache, Philippe  and
      Choukri, Khalid  and
      Cieri, Christopher  and
      Declerck, Thierry  and
      Goggi, Sara  and
      Isahara, Hitoshi  and
      Maegaard, Bente  and
      Mariani, Joseph  and
      Mazo, H{\'e}l{\`e}ne  and
      Moreno, Asuncion  and
      Odijk, Jan  and
      Piperidis, Stelios",
    booktitle = "Proceedings of the Twelfth Language Resources and Evaluation Conference",
    month = may,
    year = "2020",
    address = "Marseille, France",
    publisher = "European Language Resources Association",
    url = "https://aclanthology.org/2020.lrec-1.402",
    pages = "3291--3296",
    language = "English",
    ISBN = "979-10-95546-34-4",
}

@article{penedo2025fineweb2,
  title={FineWeb2: One Pipeline to Scale Them All--Adapting Pre-Training Data Processing to Every Language},
  author={Penedo, Guilherme and Kydl{\'\i}{\v{c}}ek, Hynek and Sabol{\v{c}}ec, Vinko and Messmer, Bettina and Foroutan, Negar and Kargaran, Amir Hossein and Raffel, Colin and Jaggi, Martin and Von Werra, Leandro and Wolf, Thomas},
  journal={COLM},
  year={2025}
}

@misc{apertus,
      title={Apertus: Democratizing Open and Compliant LLMs for Global Language Environments}, 
      author={Alejandro Hernández-Cano and Alexander Hägele and Allen Hao Huang and Angelika Romanou and Antoni-Joan Solergibert and Barna Pasztor and Bettina Messmer and Dhia Garbaya and Eduard Frank Ďurech and Ido Hakimi and Juan García Giraldo and Mete Ismayilzada and Negar Foroutan and Skander Moalla and Tiancheng Chen and Vinko Sabolčec and Yixuan Xu and Michael Aerni and Badr AlKhamissi and Ines Altemir Marinas and Mohammad Hossein Amani and Matin Ansaripour and Ilia Badanin and Harold Benoit and Emanuela Boros and Nicholas Browning and Fabian Bösch and Maximilian Böther and Niklas Canova and Camille Challier and Clement Charmillot and Jonathan Coles and Jan Deriu and Arnout Devos and Lukas Drescher and Daniil Dzenhaliou and Maud Ehrmann and Dongyang Fan and Simin Fan and Silin Gao and Miguel Gila and María Grandury and Diba Hashemi and Alexander Hoyle and Jiaming Jiang and Mark Klein and Andrei Kucharavy and Anastasiia Kucherenko and Frederike Lübeck and Roman Machacek and Theofilos Manitaras and Andreas Marfurt and Kyle Matoba and Simon Matrenok and Henrique Mendoncça and Fawzi Roberto Mohamed and Syrielle Montariol and Luca Mouchel and Sven Najem-Meyer and Jingwei Ni and Gennaro Oliva and Matteo Pagliardini and Elia Palme and Andrei Panferov and Léo Paoletti and Marco Passerini and Ivan Pavlov and Auguste Poiroux and Kaustubh Ponkshe and Nathan Ranchin and Javi Rando and Mathieu Sauser and Jakhongir Saydaliev and Muhammad Ali Sayfiddinov and Marian Schneider and Stefano Schuppli and Marco Scialanga and Andrei Semenov and Kumar Shridhar and Raghav Singhal and Anna Sotnikova and Alexander Sternfeld and Ayush Kumar Tarun and Paul Teiletche and Jannis Vamvas and Xiaozhe Yao and Hao Zhao Alexander Ilic and Ana Klimovic and Andreas Krause and Caglar Gulcehre and David Rosenthal and Elliott Ash and Florian Tramèr and Joost VandeVondele and Livio Veraldi and Martin Rajman and Thomas Schulthess and Torsten Hoefler and Antoine Bosselut and Martin Jaggi and Imanol Schlag},
      year={2025},
      eprint={2509.14233},
      archivePrefix={arXiv},
      primaryClass={cs.CL},
      url={https://arxiv.org/abs/2509.14233}, 
}

@book{OECD2013,
  title = {OECD skills outlook 2013: first results from the survey of adult skills},
  author = {Organisation for Economic Co-Operation and Development},
  year = {2013},
  publisher = {Organization for Economic Co-operation and Development (OECD)},
  address = {Paris Cedex, France},
}

@article{Stajner2022,
  title = {Lexical simplification benchmarks for English, Portuguese, and Spanish},
  author = {Štajner, Sanja and Ferrés, Daniel and Shardlow, Matthew and North, Kat and Zampieri, Marcos and Saggion, Horacio},
  journal = {Frontiers in Artificial Intelligence},
  volume = {5},
  pages = {991242},
  year = {2022},
}

@book{OECD2020,
  title = {Improving educational equity in Romania},
  author = {Organisation for Economic Co-operation and Development (OECD)},
  year = {2020},
  doi = {10.1787/f4a8c506-en},
}

@article{Coleman2021,
  title = {Preparing accessible and understandable clinical research participant information leaflets and consent forms: a set of guidelines from an expert consensus conference},
  author = {Coleman, E. and others},
  journal = {Research Involvement and Engagement},
  volume = {7},
  number = {1},
  pages = {31},
  year = {2021},
}

@article{Kim2022,
  title = {Health Literacy Level and Comprehension of Prescription and Nonprescription Drug Information},
  author = {Kim, M. and Suh, D. and Barone, J. A. and Jung, S.-Y. and Wu, W. and Suh, D.-C.},
  journal = {International Journal of Environmental Research and Public Health},
  volume = {19},
  number = {11},
  year = {2022},
  doi = {10.3390/ijerph19116665},
}

\appendix

\section{DexFlex Pipeline}
\label{sec:dexapp}
The pipeline used for this study contains several stages which are detailed below.

The first step in suggesting synonyms using the DexFlex framework is identifying the target word along with a series of additional information about it. More specifically, we use spaCy to find the current part of speech of the word and to get the necessary details regarding various grammatical attributes such as number, person, or gender.

The second step in the pipeline involves selecting synonyms from the dexonline database. For better accuracy of this process, we first establish the contextual meaning of the word by comparing the current sentence context with representative contextual examples of the alternatives found in the dexonline database. The contextual examples found in the dexonline database are stored as cached embeddings and aproximate nearest neighbour search is used to identify the meaning with the highest cosine similarity. The synonyms are retrieved as lemmas. Summary evaluations showed that this approach is reliable in correctly disambiguating words that have multiple meanings for Romanian, but we did not run exhaustive word-sense-disambiguation evaluation. 

The third step in the pipeline involves bringing the substitution candidates to the correct inflected form. In this regard, we use the grammatical knowledge derived from spaCy \texttt{ro\_core\_news\_lg} together with simple pre-defined rules to retrieve inflected forms from the dexonline database.

\section{Simplicity Ranking from Pair-wise Assessments}
\label{app:pair}
We use the methodology proposed by \citet{jerdee2023luck} to estimate a ranking from pair-wise binary simplicity scores assigned by annotators to sentences.

Considering a set of $n$ replacement candidates labeled by $i = 1 \cdot n$, assign to each a real score parameter $s_i \in [-\infty, \infty]$. Then the probability that $i$ is simpler than $j$ is assumed to be some function of the difference of their scores: $p_{ij} = f(s_i - s_j)$. The function $f(s)$ satisfies the following axioms: it is increasing in $s$, it tends to $1$ as $s \rightarrow \infty$ and to $0$ as $s \rightarrow -\infty$, and it is asymmetric about its mid-point at $s = 0$ with the form $f(-s) = 1 - f(s)$. The logistic function is a popular choice, which gives $f(s_i - s_j) = \frac{e^{s_i}}{e^{s_i}+e^{s_j}}$ also known as the Bradley-Terry model.

Now, suppose we observe $m$ matches between $n$ players. The outcomes of the matches can be represented by an $n \times n$ matrix $A$ with element $A_{ij}$ equal to the number of times player $i$ beat player $j$. The probability of the complete set of observed outcomes is $P(A|s) = \displaystyle\prod_{ij} f(s_i - s_j)^{A_{ij} } = \displaystyle\prod_{ij} \left(\frac{e^{s_i}}{e^{s_i}+e^{s_j}} \right) ^{A_{ij}}$, where $s$ is the vector with elements $s_i$.

We calculate a maximum a posteriori (MAP) estimate of the values of the scores as: $\hat{s} = argmax_s P(s|A) = argmax_s P(A|s)P(s)$, given a prior with the variance chosen as $\frac{1}{2}$: $P(s) = \displaystyle\prod_{i=1}^{n} \frac{1}{\sqrt{\pi}}e^{-s_i^2}$. According to \citet{jerdee2023luck}, the MAP estimate always exists regardless of whether the interaction network is strongly connected or not and using a prior eliminates the need for normalization.

\section{Literacy in Romania}
\label{app:literacy}

According to the adult literacy report conducted in 24 highly-developed countries \cite{OECD2013,Stajner2022}, 16.7\% of the population, on average, cannot understand texts that go beyond a basic vocabulary. Furthermore, functional literacy in Romania remains among the lowest in the European Union, with around 40\% of students not achieving baseline proficiency in at least one subject in PISA 2015 \cite{OECD2020}.

Further research has correlated low literacy with health risks - limited understanding of medication instructions \cite{Coleman2021}, misinterpretation of drug treatment information \cite{Kim2022,Nisbeth2016}, and the inability to make informed decisions in following a treatment or reading medical information. Creating text simplification systems can have several long-term societal benefits. The lack of pre-existing research and annotated data for Romanian text simplification / complexity represents an opportunity to bring novel contributions and create guidelines for developing similar approaches to new languages. 




\begin{figure}[htb]
    \centering
    \includegraphics[width=\linewidth]{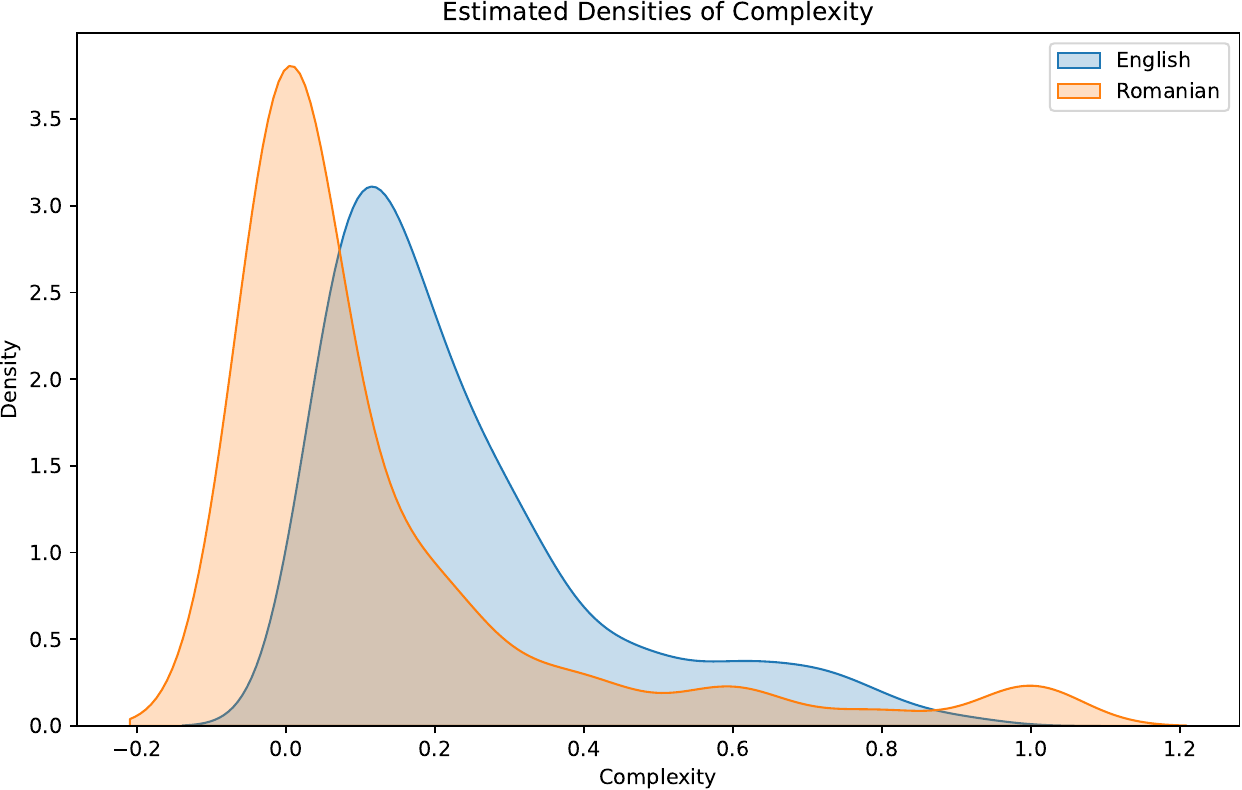} 
    \caption{Kernel density estimations of the English vs. Romanian average complexity annotations. The translation process introduces shifts in the lexical complexity. A word that is considered medium complex in English has a Romanian translation that is perceived as simpler by native Romanian speakers.}
    \label{fig:densities}
\end{figure}

\section{Romanian Original Datasets: WT and RoLCP}
\label{app:wt}
We use the Representative Corpus of Romanian \cite{midrigan-ciochina-etal-2020-resources} consisting of a diverse set (21 different genres) of written texts and speech transcripts from Romania and Moldova. The entire corpus is split into sentences using the large Romanian spaCy \cite{spacy} model. 

\begin{figure}[]
    \centering
    \includegraphics[width=\linewidth]{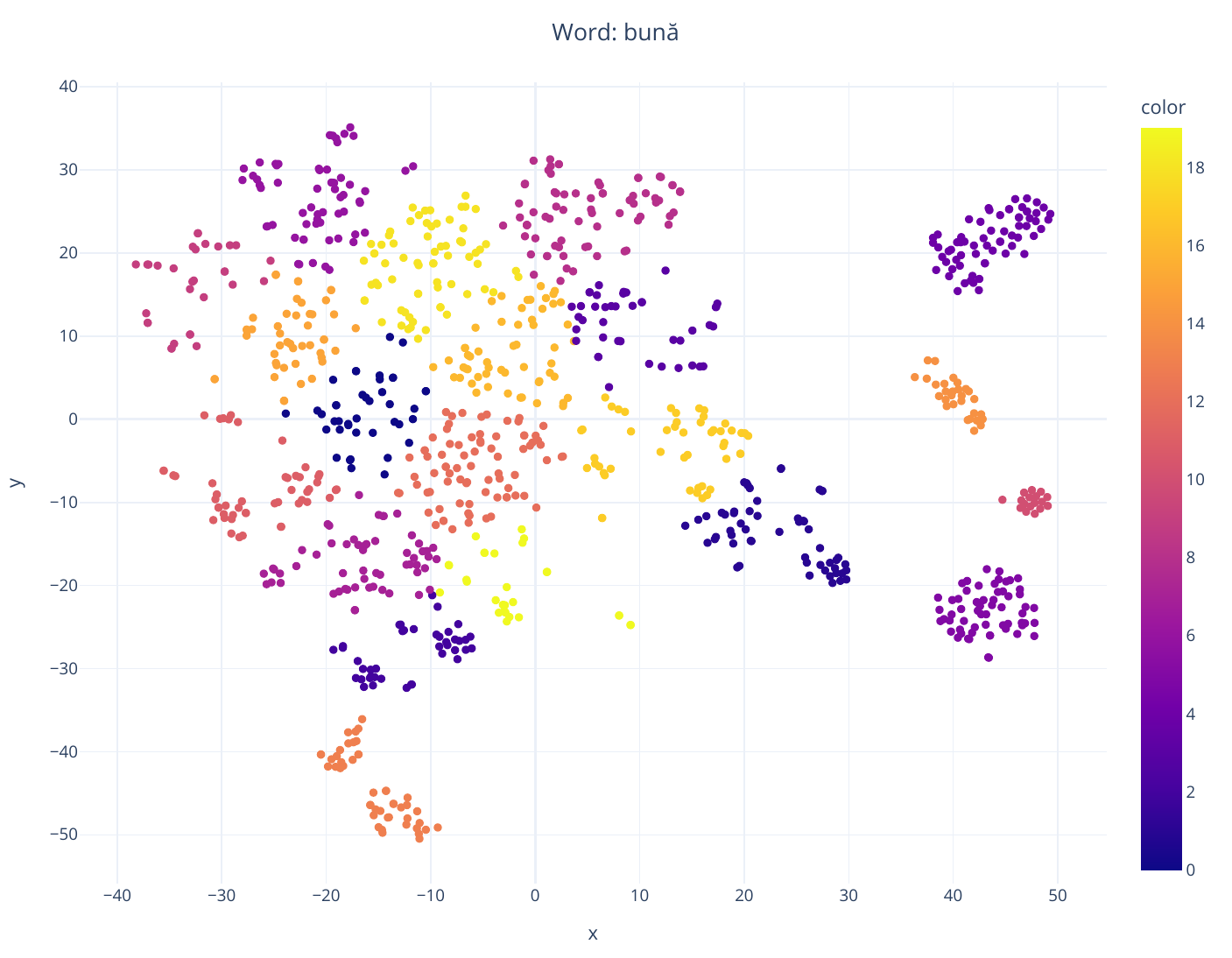} 
    \caption{The different clusters for word ``bună [en. \textit{good}]'' containing different meanings and context for the usage of the word. The clusters contain sentences with different collocations of the word (approximate translations: \textit{good day}, \textit{good decision}, \textit{good food}, \textit{good to go}, \textit{good side of things}, \textit{good will}, etc.). The sentence selection process incorporates both samples close to the centroid, representing prototypical usages, and outlier samples, reflecting less typical contexts.}
    \label{fig:wordbuna}
\end{figure}

Sentences containing each word are filtered to match the average length of our dataset. Because a word may have different meanings and functions depending on context, we apply the following procedure to construct the final sentence list. For each target word, we extract contextualized word embeddings using a Romanian BERT model \cite{dumitrescu2020birth} and project them into two dimensions with t-SNE. The resulting representations are clustered with KMeans, and the optimal number of clusters is selected according to the silhouette score. From each cluster, we sample 15\% of the sentences: half are drawn from those closest to the centroid, representing prototypical usages, and half from those farthest away, capturing peripheral or atypical contexts. This yields a balanced subset that reflects diverse instances within each cluster of meaning (an example is shown in \autoref{fig:wordbuna}). Finally, we manually review the selected samples, remove noisy sentences, and submit the remainder for explicit complexity annotation. The final dataset contains 1,765 sentences, with statistics summarized in \autoref{tab:stats}. Its mean complexity is slightly higher than that of the English and human-translated datasets, and the difference is statistically significant ($p < 0.001$) according to a bootstrapping permutation test.

We construct a third set of annotations (dataset RoLCP) using sentences sourced from original Romanian texts, without constraints imposed by a predefined list of target words. To this end, we select 12 texts spanning diverse genres, including Wikipedia articles, popular science, literature, institutional documents, and argumentative essays. Each text is sentence-split, and up to five target words per sentence are selected for annotation. The selection of target words is based on precomputed frequency statistics from a large Romanian corpus derived from \cite{wordfreq}, so that annotators assess both high-frequency and low-frequency words. 

The pool of annotators have been randomly assigned different samples, yielding a total of 10 annotations per sample. The annotation process takes place in a lab in complete silence, annotators have been given a practice dataset of 15 sentences before beginning the actual process. During the lab-centered annotation process, annotators may ask questions and clarify corner cases with the supervisors.

\section{Lexical Simplification Prompt}
\label{app:prompt}

\texttt{Provide a list of 10 alternative simpler words (as a json object) that a child would understand easily to replace the word "ORIGINAL\_WORD" in the context of the following sentence. It is mandatory to use suitable meanings for the context of the sentence and for the pattern of the answer to be displayed as a JSON with words as keys and complexity scores as values with all the 10 alternatives. Provide only words in "LANGUAGE". Sentence: "ORIGINAL".}

\end{document}